\documentclass[preprint,12pt, a4paper]{elsarticle}

\usepackage{amssymb}

\usepackage{lineno}
\usepackage{hyperref}
\usepackage[utf8]{inputenc}
\usepackage{graphicx,url}
\usepackage{float,amssymb}
\usepackage{mathptmx,algorithmicx}
\usepackage[ruled,vlined]{algorithm2e}
\usepackage{subfig}
\usepackage{float}
\restylefloat{table}

\usepackage{amsmath}
\usepackage{threeparttable}

\journal{SoftwareX}

\begin{document}

\begin{frontmatter}



\title{GSGP-CUDA - a CUDA framework for Geometric Semantic Genetic Programming}


\author{Leonardo Trujillo$^{1}$}

\author{Jose Manuel Muñoz Contreras$^{1}$}

\author{Daniel E Hernandez$^{1}$\corref{cor1}}

\author{Mauro Castelli$^2$}

\author{Juan J Tapia$^3$}

\address{$^{1}$ Tecnol\'ogico Nacional de M\'exico/IT de Tijuana,  Tijuana, B.C., Mexico \\
$^{2}$ NOVA Information Management School (NOVA IMS), Universidade Nova de Lisboa, Campus de Campolide, 1070-312, Lisboa, Portugal \\
$^{3}$ Instituto Polit\'ecnico Nacional - CITEDI, Av. Instituto Polit\'ecnico Nacional No. 1310 Colonia Nueva Tijuana, C.P.
22435, Tijuana, B.C., Mexico}

\cortext[cor1]{Corresponding author}

\begin{abstract}
Geometric Semantic Genetic Programming (GSGP) is a state-of-the-art machine learning method based on evolutionary computation. GSGP  performs search operations directly at the level of program semantics, which can be done more efficiently then operating at the syntax level like most GP systems.  Efficient implementations of GSGP in C++ exploit this fact, but not to its full potential. This paper presents GSGP-CUDA, the first CUDA implementation of GSGP and the most efficient,  which exploits the intrinsic parallelism of GSGP using GPUs. Results show speedups greater than $1,000\times$ relative to the state-of-the-art sequential implementation. 
\end{abstract}


\end{frontmatter}

\begin{table}[H]
\caption{Code metadata: General information about this code.}
\begin{tabular}{|l|p{6.5cm}|p{6.5cm}|}
\hline
\textbf{Nr.} & \textbf{Code metadata description} & \textbf{Please fill in this column} \\
\hline
C1 & Current code version & v1.0 \\
\hline
C2 & Permanent link to code/repository used for this code version & http://tiny.cc/0sbwjz$^a$\tnote{a} \\
\hline
C3 & Code Ocean compute capsule & PENDING\\
\hline
C4 & Legal Code License   & GNU General Public License v3.0 \\
\hline
C5 & Code versioning system used & None \\
\hline
C6 & Software code languages, tools, and services used & C/C++/CUDA,CUBLAS \\
\hline
C7 & Compilation requirements, operating environments \& dependencies & Toolkit CUDA v10.1 \&\& v9.2, GCC v7.4.0, CUBLAS v2.0 ,Headers Linux, unix-like systems (Ubuntu Linux18.04) \\
\hline
C8 & If available Link to developer documentation/manual & $http://PENDING/doc/$ \\
\hline
C9 & Support email for questions & leonardo.trujillo@tectijuana.edu.mx; daniel.hernandezm@tectijuana.edu.mx\\
\hline
\end{tabular}
\label{} 
\begin{tablenotes}
	\item[a]{$^a$Temporary link to source code since we experienced some difficulties with the submission system, github repository will be available publicly upon acceptance.}
\end{tablenotes}
\end{table}








\section{Motivation and significance}
\label{sec:motivation}
Geometric Semantic Genetic Programming (GSGP) is a variant of Genetic Programming (GP), a state-of-the-art machine learning algorithm
\cite{GSGP,GSGPRW,Energy,GSGPC,GSGPC2}.
GP is an evolutionary computation method that performs search and optimization following a high-level model of natural evolution.
While standard GP uses genetic operators that produce offspring by manipulating the syntax of the parents,
GSGP performs search operations directly on semantic space,
which induces a unimodal fitness landscape that simplifies the search \cite{GSGPLS}.

In symbolic regression problems, the application domain for GSGP-CUDA,
semantic space is $\mathbb{S}\subseteq \mathbb{R}^n$,
where $n$ is the number of training samples or fitness cases, allowing the search operators to be implemented more efficiently than syntax manipulations.
This allowed Vanneschi et al. \cite{GSGPRW} to implement an efficient version of GSGP using C++ pointers \cite{GSGPC};
this implementation is the reference baseline for the current work and is hereafter
referred to as GSGP-C++. A newer version of this library has been published \cite{GSGPC2},
but the core search algorithm remains unchanged.

GSGP-CUDA is a GSGP framework that exploits the parallel nature of GSGP using Graphical Processing Units (GPUs).
The goal is to allow GSGP practitioners to efficiently tackle large problems.
Search operators in GSGP define linear combinations of parent individuals, such that the semantic vector
of an offspring is defined as a linear combination of the semantic vectors of the parent(s).
This operation is naively parallel, allowing for a direct implementation in GPUs.
Such architectures have been exploited in deep learning \cite{handson}, but have been sparsely used with GP \cite{Chitty2016FasterGP, InterperterLangdom, Langdon2013}. Note that the works presented by Chitty \cite{Chitty2016FasterGP}, and Langdon and Banzhaf \cite{InterperterLangdom, Langdon2013} focus on implementing CUDA and other parallel tools to accelerate the interpretation of syntax trees, since that is the representation most commonly used in GP. Nevertheless, while GSGP can use syntax trees as well, this is not a requirement of the main evolutionary loop, particularly since syntax is only evaluated for the initial population. In the present work, for example, a linear representation is used instead. GSGP creates new individuals through the linear combination of the semantic vector of the parents, instead of combining or modifying syntax, as traditional GP algorithms do. The main issue has been that syntactic manipulation of individuals is not as efficient in GPUs, compared to numerical computations. With GSGP, on the other hand, the synergy with modern GPU architectures is stronger.

\section{Software description}
\label{sec:description}
GSGP can be briefly summarized as follows \cite{GSGP,GSGPIntro}, focusing on
the most common application domain, real-valued symbolic regression.
In this problem, fitness cases (or dataset samples) are defined as $\mathbf{fc}_j= (\mathbf{x}_j,t_j)$
with $t_j \in \mathbb{R}$, $\mathbf{x}_j \in \mathbb{R}^l$  and $l$ is the number of input features.
In standard GP and GSGP individuals are usually represented using syntax trees, so we refer to individuals as programs or trees w.l.o.g.
The semantics $\mathbf{s}_{T_i} =\{s_{T_i,1}, ..., s_{T_i,n}\}$ of a tree $T_i$ is a vector of size $n$, which represents the total
number of fitness cases, where $s_{T_i,j}$ is the output of the \textit{i-th} tree evaluated on the inputs from the \textit{j-th} fitness case.
In this way, the fitness of a tree $T_i$ is computed as the distance between $\mathbf{s}_{T_i}$ and the target semantics $\mathbf{t} = [t_1, ..., t_n]$.
While the original GSGP formulation included both geometric semantic crossover (GSC) and mutation (GSM) operators,
experimental work suggests that mutation is usually sufficient to perform an evolutionary search \cite{GSGPRW}.
A mutated offspring of parent $T$ is defined as $T_M = T + ms\cdot (R_1 + R_2)$ with $R_u, u=1,2$ are random real functions with range $[0,1]$,
constructed as $R_u = (1+ e^{-T_u})^{-1}$ with $T_u$ is a random GP tree, and $ms$ is the mutation step.
A useful property of both GSM and GSC is that the semantics of an offspring is basically defined by a linear combination of the semantics of the parents and the random trees.
Therefore, it is possible to compute the semantics and fitness of an offspring using only the semantic vectors.
This allows for an efficient implementation of the GSGP search, by building and updating semantic matrices $S_{m\times n}$, where $m$ is the number
of individuals (in either the population or the set of random trees), as done in GSGP-C++ \cite{GSGPC,GSGPC2}.

\subsection{Software Architecture}
\label{sec:architecture}
CUDA has emerged as a general purpose parallel computing API for NVDIA GPUs, designed to work
with high-level programming languages (C, C++ and Fortran).
CUDA is the most widely used computing platform for general purpose High Performance Computing (HPC).

\begin{figure}[t]
\centering
\includegraphics[width=0.75\textwidth]{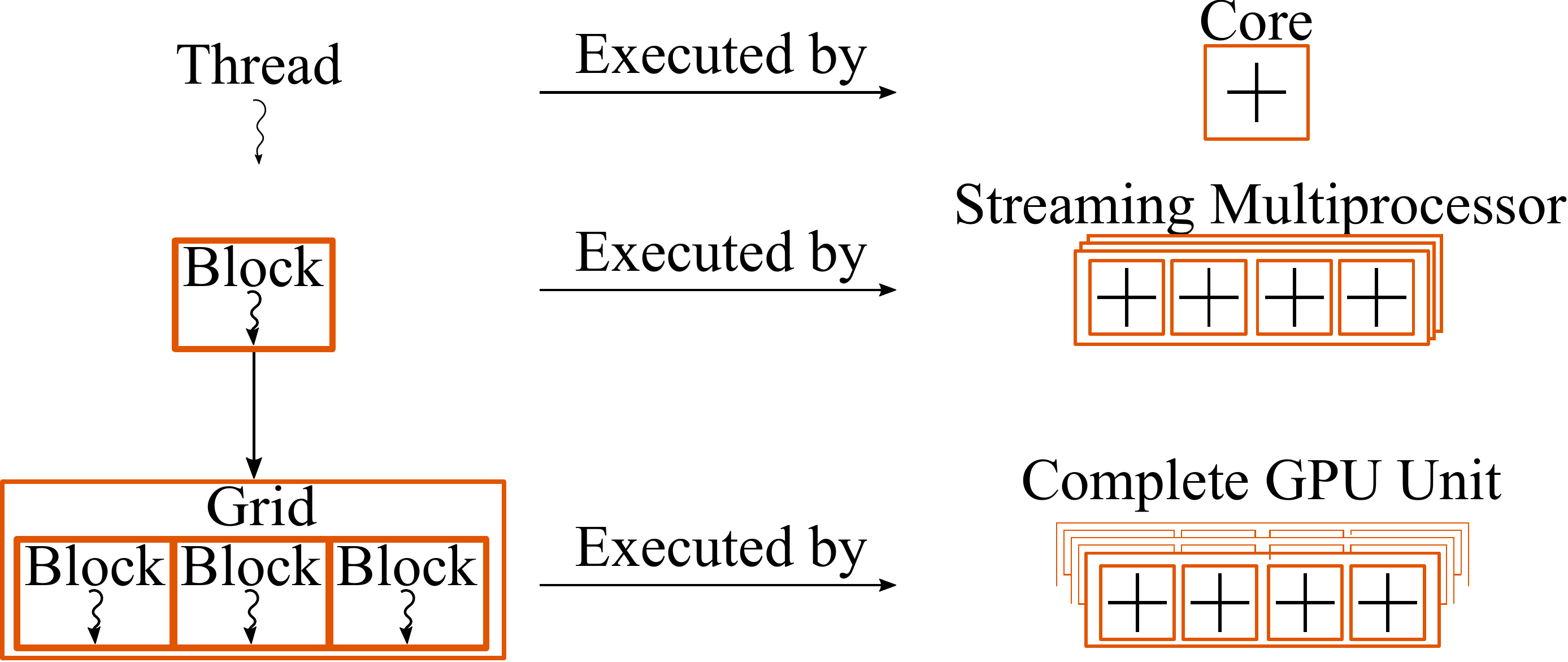}
\caption{GPU execution model.}
\label{cuda:execution} 
\end{figure}

CUDA programming is heterogeneous since both the GPU (device) and CPU (host) can be used.
The code on the CPU manages memory and launches kernels that are executed on the GPU.
Parallelism arises in the execution of kernels, which are launched over multiple threads simultaneously. 
The CUDA execution model relies on three main concepts:
thread, blocks and grids; as shown in Figure \ref{cuda:execution}.


GSGP-C++ manages the evolutionary cycle of GSGP using semantic matrices
therefore efficiently migrating these structures to a GPU is possible.
While previous works have developed GPU-based implementations of GP \cite{Robilliard2009HighPG,InterperterLangdom,Harding2011DistributedGP,AUGUSTO201386},
manipulating syntax trees is not efficient.
However, since GSGP does not require explicit construction of program syntax, it can exploit
efficient matrix manipulation on GPUs.
The evolutionary process in GSGP-CUDA has the following stages: 

\begin{enumerate}
    \item Calculating ranges of data and process parallelization. Depending on the problem size (features $l$, fitness cases $n$)
    and GSGP configuration (population size $m$ and maximum program size $k$),
    it is necessary to define the configuration of parallel work at the thread, block and grid level. 
    
    \item Creation of memory variables. All global memory regions must be reserved and allocated on the device.
    The manipulation of data is through pointer vectors in each stage.
    
    \item Initialization. Creates the initial population on the GPU, using a linear genome representation
    for the programs \cite{InterperterLangdom,Langdon2013}.
    Structures are used to keep track of the evolutionary lineage of each individual.
    
    \item Fitness evaluation. There are two kernels used in this stage. The first
    kernel is an interpreter that computes the semantics of each individual based on their genome (and their offspring),
    while the second kernel computes the fitness given by the Root Mean Squared Error (RMSE).
    
    \item GSGP Evolution. This process includes applying GSM on the parent population to compute the offspring semantics and
    the offspring fitness, and performing survival for the next generation.
    Since GSGP-CUDA is intended for large problems, selection pressure is kept low by
    applying mutation to all the individuals in the population \cite{GSGP}.
\end{enumerate}

\subsection{Software Functionalities}
\label{sec:function}
The main evolutionary algorithm in GSGP-CUDA is controlled by the CPU with a sequential flow,
which launches each of the CUDA kernels that 
perform each of the evolutionary processes using Single Instruction Multiple Data (SIMD) execution, summarized in Algorithm \ref{Alg:gsgpMain}.
For each kernel, Algorithm \ref{Alg:gsgpMain} specifies the level of parallelism that was implemented.
This refers to what is being executed on an individual CUDA thread.
Some parameters are based on the characteristics of the GPU card.
Blocks and grids are three-dimensional structures of size
$(threads.x, threads.y,threads.z)$ and $(blocks.x, blocks.y, blocks.z)$ respectively.
In this implementation grids and blocks are one-dimensional so only $threads.x$ and $blocks.x$ are considered,
with the other dimensions set to $1$.
Therefore, we will refer to variable $threads$ and $blocks$ to define the size of these structures.
The $blocks$ and $threads$ variables are set using the CUDA Occupancy API at run time.
This is done using the CUDA Occupancy API (available in CUDA 6.5 and later), namely the \textit{cudaOccupancyMaxPotentialBlockSize}
function that heuristically calculates a block size that achieves the maximum multiprocessor-level occupancy for each kernel.

\begin{algorithm}[t]
\caption{GSGP-CUDA Evolutionary Algorithm}
\footnotesize
\label{Alg:gsgpMain}
\textbf{Inputs:} Population size $m$, number of random trees $r$, fitness cases ($n\times l$), target semantics $\mathbf{t}$, maximum program size $k$ and number of generations $g$;\\
\textit{kernel} CreatePopulation($m,k$) $\rightarrow P$; \Comment{Parallelism at the gene level} \\
\textit{kernel} CreatePopulation($r,k$) $\rightarrow R$; \Comment{Random Trees used for GSM}\\
\textit{kernel} ComputeSemantics($P$) $\rightarrow S_{T,m\times n}$; \Comment{Parallelism at the individual level}\\
\textit{kernel} ComputeSemantics($R$) $\rightarrow S_{R,r\times n}$; \Comment{Semantic matrix of the random trees}\\
\textit{kernel} ComputeFitness($S_{T,m\times n},\mathbf{t}$) $\rightarrow F_{T,m\times 1}$; \Comment{Parallelism at the individual level}\\
\nl
\For {generation $<=$ to $g$ }{
\textit{kernel} GSM($ S_{T,m\times n},  S_{R,r\times n}$) $\rightarrow S_{O,m\times n}$; \Comment{Parallelism at the semantic element level}\\
\textit{kernel} ComputeFitness($S_{O,n\times m},\mathbf{t}$) $\rightarrow F_{O,m\times 1}$; \Comment{Fitness of the offspring} \\
Survival with elitism; \Comment{Implemented using CUBLAS and pointer aliasing} \\
}
\textbf{Return:} Best individual found; 
\end{algorithm}

\paragraph{Initialization}
The \textit{CreatePopulation} kernel creates the population of programs $T$ and the set of random trees $R$ used by the GSM kernel,
based on the desired population size and maximum program length.
The individuals are represented using a linear genome, composed of valid terminals (inputs to the program) and functions (basic elements with which
programs can be built).
Terminals include input features and ephemeral random constants, while a reduced function set of arithmetic operators is used as suggested in \cite{GSGPRW},
of $+,-,\times$ and protected division $\div$.
Each gene in the chromosome is determined randomly based on the following probabilities:
$0.8$ for a function, $0.14$ for a problem variable or feature, and $0.04$ for an ephemeral random constant.

\paragraph{Interpreter}
The \textit{ComputeSemantics} kernel is basically an interpreter, that decodes each GP tree and
evaluates it over all the fitness cases, producing as output the semantic vector of each individual.
The chromosome is interpreted linearly, using an auxiliary LIFO stack $D$ that stores terminals from the chromosome
and the output from valid operations, following \cite{push}.
When a terminal gene in a chromosome is encountered by the interpreter, a $push$ of the value into stack $D$ is done.
Conversely, when a function gene is encountered, $pop$ operations are performed on $D$ to obtain the arguments (operands) for the function;
the number of $pop$ operations is equal to the arity of the function.
After computing the function, the output is pushed onto the stack $D$.
If the number of elements in $D$ is less then the arity of the function, then the operation is skipped and the stacked is left
in the original state (before encountering the function gene in the chromosome).
An additional output register $Out$ is used to store all the valid outputs from computed functions.
When the interpreter gets to the end of the chromosome, the kernel returns the contents of $Out$.
This is repeated for each fitness case in the training dataset.

Each individual is executed over a single thread, over all the fitness cases.
The \textit{ComputeSemantics} kernel produces the semantic matrix of the population $S_{T,m\times n}$ and the semantic matrix
of the random trees $S_{R,r\times n}$.

\paragraph{Fitness computation}
The \textit{ComputeFitness} kernel computes the RMSE between each row of the semantic matrix $S_{T,m\times n}$ and the target vector $\mathbf{t}$, computing the fitness of each individual in the population.
To account for large datasets with many fitness cases, and large populations, an efficient partition of the
semantic matrix is necessary.
Each thread computes the fitness of a single individual.

\paragraph{Geometric Semantic Mutation}
The search operation is performed by the \textit{GSM} kernel.
The GSM operator is basically a vector addition operation, that can be performed independently for each semantic element $s_{T_i,j}$.
However, it is necessary to select the semantics of two random trees $R_u$ and $R_v$, and a random mutation step $ms$.
In other words, while the mutation operator is defined at the individual level, it can be efficiently implemented at the semantic element level,
where each thread computes a particular semantic element in the offspring.
Therefore, to synchronize this operation over multiple threads, an array of indices and an array of mutation steps is first constructed with two auxiliary kernels,
the first one stores the indices $u$ and $v$ of the random trees used in each mutation operation while the latter stores the corresponding
mutations steps $ms$.
Afterward, the \textit{GSM} kernel uses these arrays to compute each element of the offspring semantic vector.

\paragraph{Survival}
Survival uses a generational strategy, with elitism of the best individual in both the parent and offspring populations.
This stage replaces the parent population,
semantic matrix and fitness vector with those of the offspring.
It relies on the CUBLAS ISAMIN function to identify the best individual.
To find the worst individual to be replaced, in case that the elite individual is in the parent population, the CUBLAS ISAMAX function is used.

\subsection{Complexity Analysis}
According to \cite{GSGPIntro}, the complexity of the sequential GSGP depends on the population size $(m)$ and number of generations $(g)$
resulting in $O(mg)$ time complexity; since the computational cost of evaluating a new individuals at each generation is constant $O(1)$.
However, in general, this evaluation also depends on the number of fitness cases $n$, giving a complexity of $O(mgn)$.
The proposed implementation has a complexity of $O(\dfrac{mgn}{t})$, where $t$ is the number of threads.
For modern GPUs, $t$ could approximate $m$, $g$ or $n$ in many cases, such that the complexity could be reduced by one factor.

Similarly, the memory complexity depends on the population size $(m)$ and the number of fitness cases $(n)$, but since the same arrays are used in all the iterations, the number of generations $(g)$ does not affect the memory complexity. Nevertheless, there are two additional dimensions that impact the memory complexity, these are: the maximum program size $(k)$ and the number of random trees $(r)$. Note that the first one is only used on the array for the syntax of the individuals in the initial population, and the latter is constant throughout the whole evolutionary process. 

In this way, for the initialization stage, the algorithm implements an array of floats of size $O(mk)$ for the initial population, another one of size $O(rk)$ for the random trees used for the mutation operation, one more of size $O(mn)$ that holds the semantic vector for each individual, another of size $O(rn)$ for the semantic vectors of the random trees; and finally, one of size $O(nl)$ for the input data, given by the number of fitness cases and the number of features in the problem. For the remaining iterations of the algorithm, an additional array is used to store the semantic vectors of the new individuals created at each iteration, this array is of size $O(mn)$. Hence, the memory complexity of the proposed algorithm is $O(mk+rk+2mn+rn+nl)$. As it will be described in the next section, the dimensions that present the largest values are the number of fitness cases $(n)$ and the population size $(m)$. Thus, the memory growth is described by $O(2mn)$.        

\section{Illustrative Examples}
\label{sec:examples}
The experimental evaluation has two goals.
First, to measure the speedup of GSGP-CUDA relative to GSGP-C++.
Second, to compare the performance of both implementations on real-world problems.

\subsection{Performance speedup}
The algorithms are executed on the following hardware.
For GSGP-C++ we use a Dell PowerEdge R430 server with Intel(R) Xeon(R) CPU E5-2650 v4 @ 2.20 GHz processors,
using virtual machines with 8 Xeon Cores and 32 GB of RAM running on a
Kernel-based Virtual Machine server.
For GSGP-CUDA we use two devices, an NVIDIA Quadro P4000 and an NVIDIA Tesla P100; see Table \ref{tab:DeviceSpecifications}.
The Quadro card works on a Dell precision 7811 workstation with an Intel (R) Xeon CPU E5-2603 v4 1.70Ghz and 32 GB of RAM with 64-bit Ubuntu Linux.
The Tesla card runs on IBM Power8 S822LC server for HPC,
with two Power8NVL 4.20 GHz processors with 512 GB of RAM and Ubuntu Server 16.04.

\begin{table}[t]
\caption{Specifications of GPU devices.}
\footnotesize
\label{tab:DeviceSpecifications}
\begin{tabular}{@{}lllll@{}}
\hline
\textbf{Features}     &  & \textbf{Quadro P4000}              &  &   \textbf{Tesla P100} \\ 
\hline
GPU Architecture      &  & NVIDIA Pascal       &  & NVIDIA Pascal                                                                                          \\
GPU Memory            &  & 8 GB GDDR5          &  & \begin{tabular}[c]{@{}l@{}}16GB CoWoS HBM2 at 732 GB/s or\\ 12GB CoWoS HBM2 at 549 GB/s\end{tabular} \\
Memory Interface      &  & 256-bit             &  & 4096-bit                                                                                               \\
Memory Bandwidth      &  & Up to 243 GB/s      &  & 732 GB/s or 549 GB/s                                                                                   \\
NVIDIA CUDA ® Cores   &  & 1792                &  & 3584                                                                                                   \\
System Interface      &  & PCI Express 3.0 x16 &  & NVLink                                                                                              \\
Max Power Consumption &  & 105 W               &  & 250 W                                                                                                  \\ \hline
\end{tabular}
\end{table}

To evaluate the speedup between implementations we use dummy datasets to measure execution times for different scenarios.
Table \ref{tab:param} shows the ranges and parameter values for these experiments.

\begin{table}[t]
\centering
\caption{Experimental parameters used for benchmarking GSGP-C++ and GSGP-CUDA.}
\footnotesize
\label{tab:param}
\begin{tabular}{@{}ll@{}}
\hline
\textbf{Parameter}        &   \textbf{Range}                      \\ \hline
Fitness Cases             &   $n = 102,400$         \\
Number of features        &   $l = 1,024$         \\
Program size (genes)      & $k \in [127, 1023, 2047]$ \\
Population Size           & $m = 10,240$ \\
Number of generations & $50$ \\
Runs & $30$ \\
\hline
\end{tabular}
\end{table}


The CUDA event API is used to compute the processing time, which is part of the CUDA SDK and includes calls to create and destroy events, record events and calculate the time elapsed in milliseconds between two events, using the event functions $cudaEvent\_t$, \textit{cudaEventCreate()}, \textit{cudaEventRecord()}, \textit{cudaEventSynchronize()} and \textit{cudaEventElapsedTime()}.
We compute the run times of the \textit{CreatePopulation}, \textit{ComputeSemantics} and a single GSGP generation (iteration).
In the latter case, this includes the \textit{GSM} kernel, along with all auxiliary kernels, and the survival stage.
Moreover, the total run time of a each run is computed, including load times for the training data and returning the results to the host.

Table \ref{tab:speedup} shows the speedup of GSGP-CUDA compared to GSGP-C++, for GSGP process.
Consider the case where $k=127$, which are relatively large programs and a common GP setting \cite{GSGPIntro,field}.
Notice that speedup is largest in the \textit{CreatePopulation} kernel, due to the CUDA parallelization and because
GSGP-CUDA uses a simpler linear chromosome.
In the Tesla card there is also a substantial speedup in each GSGP generation, of nearly $1,000$.
The speedup in the total run time, however, is closer to the speedup achieved in the \textit{ComputeSemantics} kernel.
This is due to two reasons.
First, \textit{ComputeSemantics} is the most computationally expensive kernel, since each individual is interpreted linearly
for each fitness case.
Second, these tests only considered 50 generations.
In many real-world problems the number of generations can be as high as $2,000$ to achieve optimal performance,
which would increase the speedup in practice.

As the individual size increases, the speedup for the $CreatePopulation$ kernel and for the GSGP Generation either stay the same or increase.
However, the speedup of the $ComputeSemantics$ kernel decreases.

\begin{table}[t]
\centering
\caption{Speed-up of GSGP-CUDA compared to GSGP++, considering populations of 10,240 individuals.}
\footnotesize
\label{tab:speedUp-2}
\begin{tabular}{@{}cccc@{}}
\hline 
\textbf{Individual size}  & \textbf{CUDA kernel} & \textbf{\begin{tabular}[c]{@{}c@{}}Speed-up\\ Quadro\end{tabular}} & \textbf{\begin{tabular}[c]{@{}c@{}}Speed-up\\ Tesla\end{tabular}} \\ \hline
127 &  \begin{tabular}[c]{@{}c@{}}CreatePopulation\\ ComputeSemantics\\ GSGP Generation\\ Total Run Time\end{tabular} & \begin{tabular}[c]{@{}c@{}} 554x
\\ 96x\\ 29x \\81x \end{tabular} & \begin{tabular}[c]{@{}c@{}} 1,177x\\ 112x\\ 872x \\ 103x\end{tabular} \\ \hline

1023 &  \begin{tabular}[c]{@{}c@{}}CreatePopulation\\ ComputeSemantics\\ GSGP Generation\\ Total Run Time\end{tabular} & \begin{tabular}[c]{@{}c@{}} 570x\\ 23x\\ 28x \\ 22x\end{tabular} & \begin{tabular}[c]{@{}c@{}} 2,039x\\ 31x\\ 850x \\ 30x\end{tabular} \\ \hline

2047 & \begin{tabular}[c]{@{}c@{}} CreatePopulation\\ ComputeSemantics\\ GSGP Generation\\ Total Run Time \end{tabular} & \begin{tabular}[c]{@{}c@{}} 1,130x\\ 24x\\ 32x \\ 23x\end{tabular} & \begin{tabular}[c]{@{}c@{}} 3,525x\\ 29x\\ 946x \\ 29x \end{tabular} \\ \hline
 &   &  &  \\
\end{tabular}
\label{tab:speedup}
\end{table}

\begin{figure}[t]
\centering
\subfloat[Quadro]{
\label{f:quadro}
\includegraphics[width=.48\textwidth]{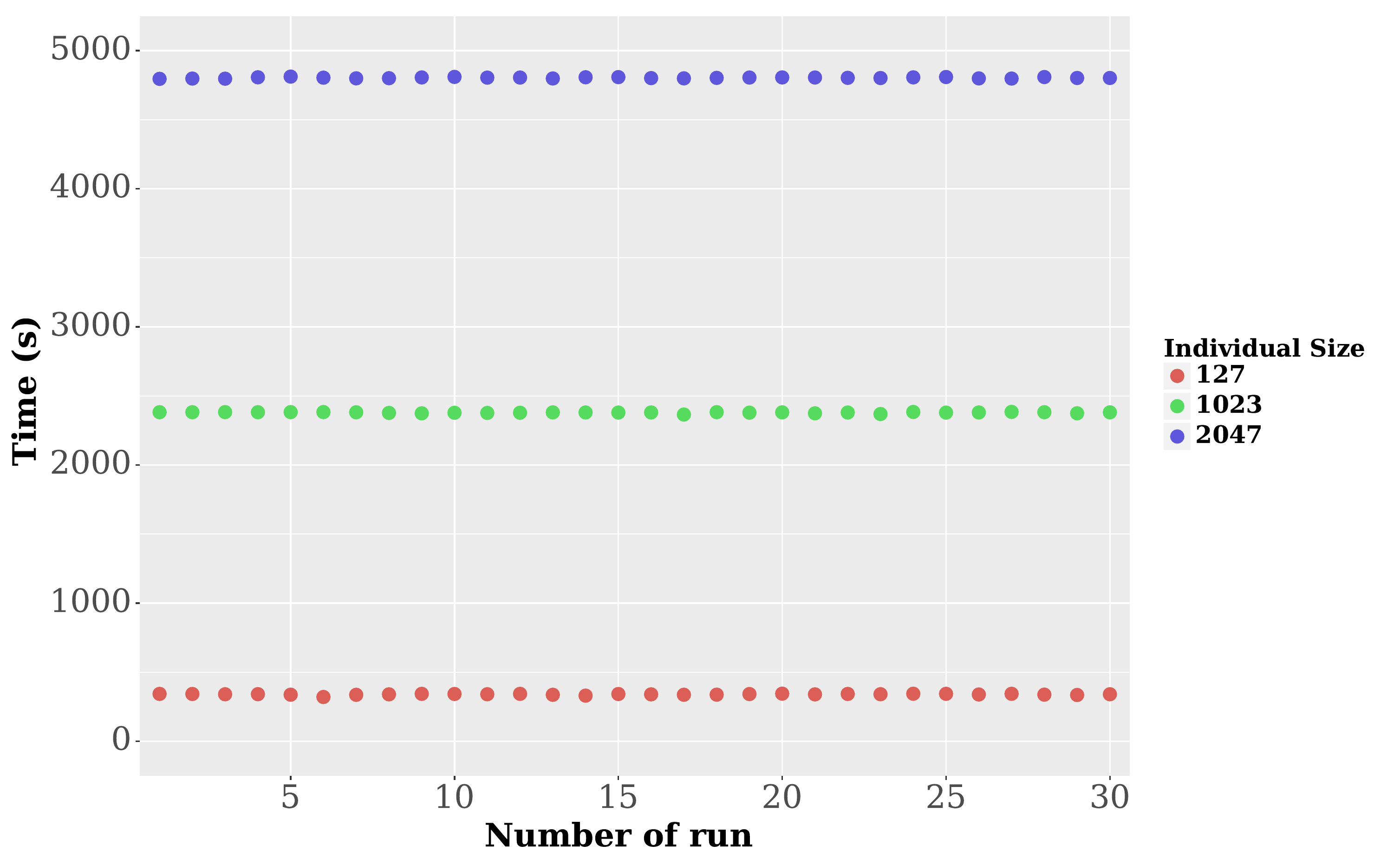}}
\subfloat[Tesla]{
\label{f:koza}
\includegraphics[width=.48\textwidth]{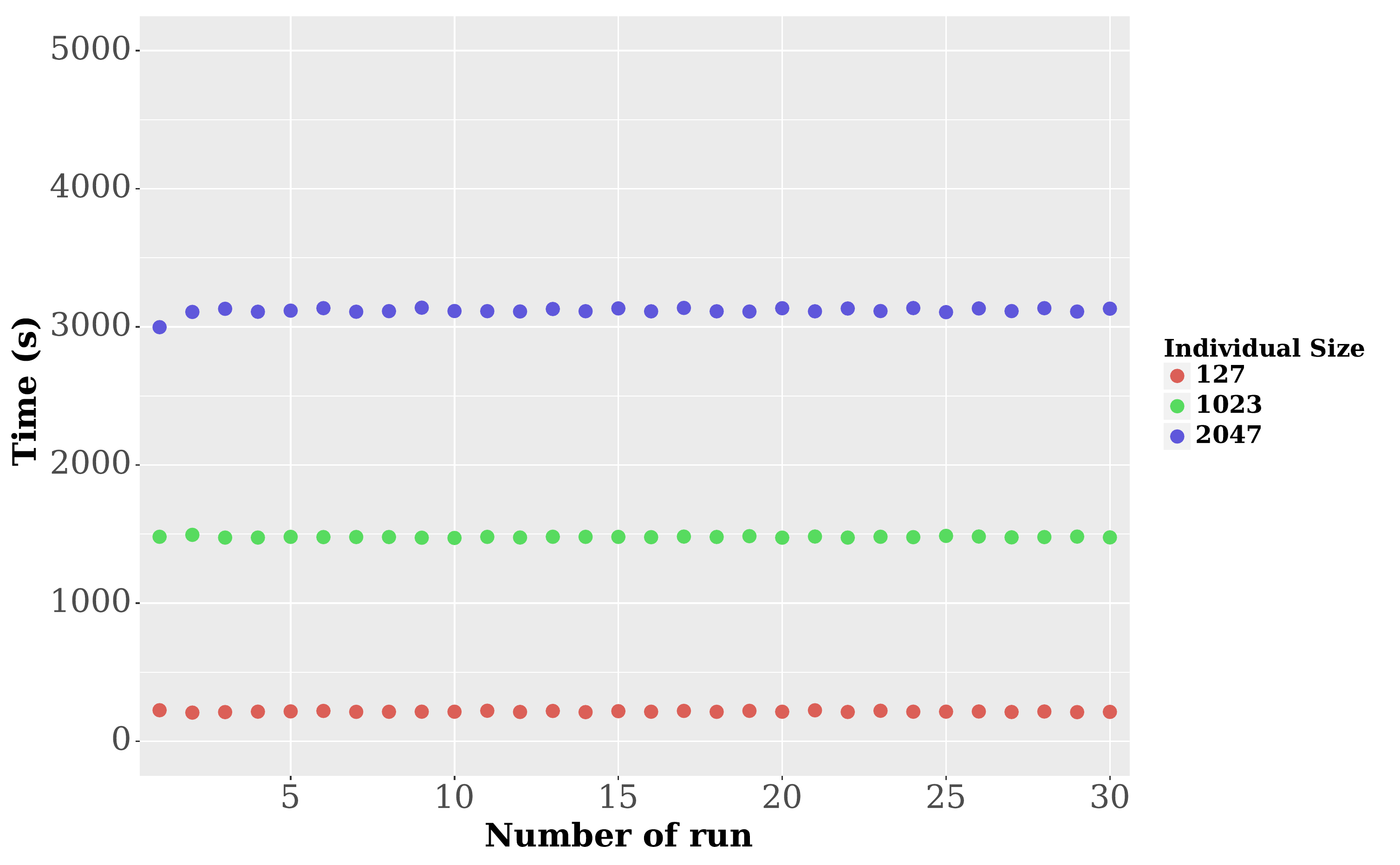}}
\caption{Performance comparison of the Quadro and Tesla cards, considering the \textit{ComputeSemantics} kernel.}
\label{fig:General_conv1}
\end{figure}

Comparing the Quadro and Tesla cards, Figure \ref{fig:General_conv1}
summarizes the results on a sweep of parameter values.
The plots show the run times of 30 runs (horizontal axis) of the \textit{ComputeSemantics} kernel,
with $102,400$ fitness cases and the Table \ref{tab:param} parameters.
Notice that the \textit{ComputeSemantics} kernel is sensitive to program size.
For these reasons, the recommended setup for GSGP-CUDA is to use small or medium-sized programs.

\subsection{Performance on real symbolic regression problems}
GSGP-CUDA and GSGP-C++ were evaluated on the set of benchmark problems in Table \ref{tab:benchmark},
which have been widely used to evaluate GP \cite{Energy,Vladislavleva2009,bench}.
The results are summarized in Table \ref{tab:fTrain} for the training performance and
in Table \ref{tab:fTest} for the testing performance, showing the mean, median, standard deviation (STD)
and interquartile range (IQR) over all the runs, after removing outliers.
Performance is comparable in most cases with only slight variations.
One notable result is that GSGP-CUDA outperforms GSGP-C++ on the most difficult problem, Tower.
This is due to the different representation used
by GSGP-CUDA.
The performance of GSGP-CUDA might be improved further, using hyperparameter optimization
or incorporating a local search method \cite{GSGPLS}.

\begin{table}[t]
\caption{Symbolic regression real world problems.}
\scriptsize
\label{tab:benchmark}
\begin{tabular}{@{}lccl@{}}
\hline
\textbf{Problems} & \textbf{No. Instances} & \textbf{No. Features} & {\textbf{Description}} \\ \hline
Concrete \cite{Yeh1998}            & 1030 & 9  & The concrete compressive strength is a highly \\
& & &                                         nonlinear function  of building parameters. \\
Energy Cooling \cite{Tsanas2012} & 768 & 9         & Assessing the cooling requirements of buildings \\
& & &                                         as a function of building parameters. \\
Energy Heating \cite{Tsanas2012}         & 768 & 9 & Assessing the heating requirements of buildings\\
& & &                                         as a function of building parameters. \\
Housing \cite{Quinlan93}              & 506  & 14 & Concerns housing values in suburbs on Boston. \\
Tower \cite{Vladislavleva2009} & 5000 & 26                & An industrial data set of chromatography measurement\\
& & &                                         of the composition of a distillation tower. \\
Yacht \cite{ortigosa2007neural} & 308 & 7                     & Delft data set, used to predict the hydrodynamic \\
& & &                                        performance of sailing from dimensions and velocity. \\ \hline
\end{tabular}
\end{table}

\begin{table}[t]
\begin{center}
\caption{Comparison of train fitness (RMSE) on the real-world benchmark problems.}
\footnotesize
\begin{tabular}{l c c c c | c c c c} 
\hline
\multicolumn{5}{c}{GSGP C++} & \multicolumn{4}{c}{GSGP CUDA} \\ \hline
Problem & Mean & Median & STD & IQR & Mean & Median & STD & IQR \\ \hline

Concrete & 8.97 & 8.87 & 0.69 & 0.46 & 8.12 & 7.89 & 0.92 & 1.65\\

Cooling & 3.50 & 3.51 & 0.24 & 0.24 & 3.25 & 3.30 & 0.18 & 0.22\\

Heating & 3.59 & 3.38 & 0.79 & 0.52 & 2.97 & 3.01 & 0.21 & 0.23\\

Housing & 4.24 & 4.27 & 0.18 & 0.26 & 4.26 & 4.25 & 0.34 & 0.32\\

Tower & 157.87 & 160.82 & 30.58 & 35.53 & 69.16 & 70.04 & 5.43 & 7.91\\

Yacht & 8.39 & 8.20 & 1.04 & 1.17 & 8.35 & 8.34 & 0.67 & 0.55\\ \hline
\end{tabular}
\label{tab:fTrain}
\end{center}
\end{table}

\begin{table}[t]
\begin{center}
\caption{Comparison of test fitness (RMSE) on the real-world benchmark problems.}
\footnotesize
\begin{tabular}{l c c c c | c c c c}
\hline
\multicolumn{5}{c}{GSGP C++} & \multicolumn{4}{c}{GSGP CUDA}\\
\hline
Problem & Mean & Median & STD & IQR & Mean & Median & STD & IQR\\
\hline
Concrete & 6.87 & 7.23 & 1.61 & 2.54 & 8.41 & 7.89 & 1.13 & 1.93\\

Cooling & 3.51 & 3.47 & 0.28 & 0.37 & 3.25 & 3.24 & 0.24 & 0.33\\

Heating & 3.67 & 3.47 & 0.81 & 0.63 & 2.99 & 3.05 & 0.23 & 0.26\\

Housing & 4.29 & 4.20 & 0.42 & 0.68 & 4.33 & 4.29 & 0.53 & 0.81\\

Tower & 156.66 & 161.16 & 26.46 & 37.87 & 69.88 & 70.89 & 5.77 & 8.75\\

Yacht & 8.62 & 8.51 & 1.10 & 1.84 & 8.79 & 8.71 & 1.14 & 1.29\\ \hline
\end{tabular}

\label{tab:fTest}
\end{center}
\end{table}

\begin{table}[t]
\centering
\caption{Experimental parameters used for benchmarking in real problems GSGP-C++ and GSGP-CUDA.}
\footnotesize
\label{tab:paramRealExperiments}
\begin{tabular}{@{}ll@{}}
\hline
\textbf{Parameter}        &   \textbf{Range}\\ \hline
Number of Runs            &   $n = 30$\\
Number of generations     &   $l = 1,024$\\
Program size (genes)      &   $k \in 1,024$\\
Population Size           &   $m = 1,024$\\
Range of ephemeral constant     & $[1,10]$\\
Probability of Mutation         & $1$\\
Probability of Crossover        & $0$\\
\hline
\end{tabular}
\end{table}

\subsection{How to use the GSGP-CUDA library}
\label{sec:5}
The run parameters are defined in a \textit{config.ini} file, namely: 
program size, population size, number of fitness cases, number of problem features, and number of runs.
To compile the gsgpCuda.cu file , simply run:\\

nvcc -std=c++11 -O0 gsgpCuda.cu -o gsgpCuda.x -lcublas\\

The program is executed with:\\

./gsgpCuda.x -train\_file train\_file.txt -test\_file test\_file.txt\\

In this case, train.txt and test.txt are the files that contain training and test instances. The columns contain the values of the problem features while the rows contain the instances. The last column contains the target variable. As output two text files are created, containing the fitness of the best individual on the training set and the same individual on the test set for each generation.

\section{Impact}
\label{sec:impact}

The main contribution of this work is to propose the first CUDA implementation of GSGP,
where the fundamental elements of the evolutionary cycle are implemented as CUDA kernels to exploit the massively parallel nature of
modern GPUs.
From a practical perspective, GSGP-CUDA is the most efficient implementation of this algorithm,
and opens up the possibility of using GSGP in particular, and GP in general, on Big Data problems.
This is important since GSGP has shown great promise on small and medium-sized problems, but previous implementations scale poorly to larger
datasets.
GP as a machine learning method has not been able to efficiently exploit modern GPU technology,
and this algorithm is a big step in this direction.
Moreover, from a scientific perspective GSGP-CUDA allows for large scale analysis of semantic space, one
of the most active research areas in GP \cite{semanticsurvey}.







\section{Conclusions}
\label{sec:conclusions}
This paper presents the first implementation of the powerful GSGP algorithm for execution on GPU cards, using CUDA programming.
Given the nature of GSGP, porting to an efficient CUDA implementation, using a linear program representation,
was demonstrated in this work.
Experimental results show that GSGP-CUDA performs similarly to the sequential implementation (GSGP-C++), but particularly performs better on the most difficult benchmark.
Moreover, benchmarking shows that the speedups of GSGP-CUDA can be as high as $1,177$x
for some process, and over $100$x for a small number of generations.
Future work will focus on integrating other search elements, such as a local search method
and extending the interpreter to allow for different data types.

\section{Conflict of Interest}
No conflict of interest exists:
We wish to confirm that there are no known conflicts of interest associated with this publication and there has been no significant financial support for this work that could have influenced its outcome.

\section*{Acknowledgements}
\label{sec:ack}
We thank Perla Ju\'arez-Smith for her help implementing some of the source code used in this software.
We also thank the Tecnol\'ogico Nacional de M\'exico/IT de Tijuana and CITEDI-IPN for their administrative
and technical assistance in the development of this work.
This research was partially supported by national funds through FCT (Fundação para a Ciência e a Tecnologia) by the projects GADgET (DSAIPA/DS/0022/2018) and AICE (DSAIPA/DS/0113/2019).
Mauro Castelli acknowledges the financial support from the Slovenian Research Agency (research core funding No. P5-0410). Funding for this work was also provided by TecNM (Mexico) 2020 through the research project ''Resoluci\'on de m\'ultiples problemas de aprendizaje supervisado de manera simult\'anea con programaci\'on gen\'etica".



\bibliographystyle{elsarticle-num} 
\bibliography{references}

\begin{thebibliography}{10}
\expandafter\ifx\csname url\endcsname\relax
  \def\url#1{\texttt{#1}}\fi
\expandafter\ifx\csname urlprefix\endcsname\relax\def\urlprefix{URL }\fi
\expandafter\ifx\csname href\endcsname\relax
  \def\href#1#2{#2} \def\path#1{#1}\fi

\bibitem{GSGP}
A.~Moraglio, K.~Krawiec, C.~G. Johnson, Geometric semantic genetic programming,
  in: Proceedings of the 12th International Conference on Parallel Problem
  Solving from Nature - Volume Part I, PPSN'12, 2012, pp. 21--31.

\bibitem{GSGPRW}
L.~Vanneschi, S.~Silva, M.~Castelli, L.~Manzoni, Geometric Semantic Genetic
  Programming for Real Life Applications, Springer New York, New York, NY,
  2014, pp. 191--209.

\bibitem{Energy}
M.~Castelli, L.~Trujillo, L.~Vanneschi, A.~Popovic, Prediction of energy
  performance of residential buildings: A genetic programming approach, Energy
  and Buildings 102 (2015) 67 -- 74.

\bibitem{GSGPC}
M.~Castelli, S.~Silva, L.~Vanneschi, A c++ framework for geometric semantic
  genetic programming, Genetic Programming and Evolvable Machines 16~(1) (2015)
  73--81.

\bibitem{GSGPC2}
M.~Castelli, L.~Manzoni, Gsgp-c++ 2.0: A geometric semantic genetic programming
  framework, SoftwareX 10 (2019) 100313.

\bibitem{GSGPLS}
M.~Castelli, L.~Trujillo, L.~Vanneschi, S.~Silva, E.~Z-Flores, P.~Legrand,
  Geometric semantic genetic programming with local search, in: Proceedings of
  the 2015 Annual Conference on Genetic and Evolutionary Computation, GECCO
  '15, 2015, pp. 999--1006.

\bibitem{handson}
A.~Gron, Hands-On Machine Learning with Scikit-Learn and TensorFlow: Concepts,
  Tools, and Techniques to Build Intelligent Systems, 1st Edition, O'Reilly
  Media, Inc., 2017.

\bibitem{Chitty2016FasterGP}
D.~M. Chitty, Faster gpu-based genetic programming using a two-dimensional
  stack, Soft Computing 21 (2016) 3859--3878.

\bibitem{InterperterLangdom}
W.~B. Langdon, W.~Banzhaf, A simd interpreter for genetic programming on gpu
  graphics cards, in: M.~O'Neill, L.~Vanneschi, S.~Gustafson, A.~I.
  Esparcia~Alc{\'a}zar, I.~De~Falco, A.~Della~Cioppa, E.~Tarantino (Eds.),
  Genetic Programming, Springer Berlin Heidelberg, Berlin, Heidelberg, 2008,
  pp. 73--85.

\bibitem{Langdon2013}
W.~B. Langdon, Large-Scale Bioinformatics Data Mining with Parallel Genetic
  Programming on Graphics Processing Units, Springer Berlin Heidelberg, Berlin,
  Heidelberg, 2013, pp. 311--347.

\bibitem{GSGPIntro}
L.~Vanneschi, An Introduction to Geometric Semantic Genetic Programming,
  Springer International Publishing, Cham, 2017, pp. 3--42.

\bibitem{Robilliard2009HighPG}
D.~Robilliard, V.~Marion, C.~Fonlupt, High performance genetic programming on
  gpu, in: BADS '09, 2009.

\bibitem{Harding2011DistributedGP}
S.~L. Harding, W.~Banzhaf, Distributed genetic programming on gpus using cuda,
  2011.

\bibitem{AUGUSTO201386}
D.~A. Augusto, H.~J.~C. Barbosa, Accelerated parallel genetic programming tree
  evaluation with opencl, J. Parallel Distrib. Comput. 73~(1) (2013) 86–100.

\bibitem{push}
L.~Spector, A.~Robinson, Genetic programming and autoconstructive evolution
  with the push programming language, Genetic Programming and Evolvable
  Machines 3~(1) (2002) 7--40.

\bibitem{field}
R.~Poli, W.~B. Langdon, N.~F. McPhee, A Field Guide to Genetic Programming,
  Lulu Enterprises, UK Ltd, 2008.

\bibitem{Vladislavleva2009}
E.~J. Vladislavleva, G.~F. Smits, D.~den Hertog, {Order of nonlinearity as a
  complexity measure for models generated by symbolic regression via pareto
  genetic programming}, IEEE Transactions on Evolutionary Computation 13~(2)
  (2009) 333--349.

\bibitem{bench}
P.~Ju{\'a}rez-Smith, L.~Trujillo, M.~Garc{\'i}a-Valdez, F.~Fern{\'a}ndez~de
  Vega, F.~Ch{\'a}vez, Local search in speciation-based bloat control for
  genetic programming, Genetic Programming and Evolvable Machines 20~(3) (2019)
  351--384.

\bibitem{Yeh1998}
I.-C. Yeh, Modeling of strength of high-performance concrete using artificial
  neural networks, Cement and Concrete Research 28~(12) (1998) 1797 -- 1808.

\bibitem{Tsanas2012}
A.~Tsanas, A.~Xifara, {Accurate quantitative estimation of energy performance
  of residential buildings using statistical machine learning tools}, Energy
  and Buildings 49 (2012) 560--567.

\bibitem{Quinlan93}
J.~R. Quinlan, Combining instance-based and model-based learning, in: Machine
  Learning, Proceedings of the Tenth International Conference, University of
  Massachusetts, Amherst, MA, USA, June 27-29, 1993, 1993, pp. 236--243.

\bibitem{ortigosa2007neural}
J.~G. I.~Ortigosa, R.~Lopez, A neural networks approach to residuary resistance
  of sailing yachts prediction., in: Proceedings of the International
  Conference on Marine Engineering MARINE, 2007, p. 250.

\bibitem{semanticsurvey}
L.~Vanneschi, M.~Castelli, S.~Silva, A survey of semantic methods in genetic
  programming, Genetic Programming and Evolvable Machines 15~(2) (2014)
  195–214.

\end{thebibliography}









\end{document}